# Application of Deep Learning on Predicting Prognosis of Acute Myeloid Leukemia with Cytogenetics, Age, and Mutations

Mei Lin, MD[1], Vanya Jaitly, MD[1], Iris Wang, MD[1], Zhihong Hu, MD[1], Lei Chen, MD[1], Md. Amer Wahed, MD[1], Zeyad Kanaan, MD[2], Adan Rios, MD[2], Andy N.D. Nguyen, MD[1*]

[1]Department of Pathology and Laboratory Medicine, University of Texas Health Science Center at Houston, Texas, TX 77030
[2]Department of Oncology, University of Texas Health Science Center at Houston, Texas, TX 77030

## Abstract

We explore how Deep Learning (DL) can be utilized to predict prognosis of acute myeloid leukemia (AML). Out of TCGA (The Cancer Genome Atlas) database, 94 AML cases are used in this study. Input data include age, 10 common cytogenetic and 23 most common mutation results; output is the prognosis (diagnosis to death, DTD). In our DL network, autoencoders are stacked to form a hierarchical DL model from which raw data are compressed and organized and high-level features are extracted. The network is written in R language and is designed to predict prognosis of AML for a given case (DTD of more than or less than 730 days). The DL network achieves an excellent accuracy of 83% in predicting prognosis. As a proof-of-concept study, our preliminary results demonstrate a practical application of DL in future practice of prognostic prediction using next-gen sequencing (NGS) data.

## INTRODUCTION

AML is a neoplasm of the bone marrow that is caused by mutations or cytogenetic (chromosomal) abnormalities in the myeloid stem cells leading to the formation of aberrant myeloblasts.[1] The highly proliferative cancer cells impede the formation of normal blood cells; leading to death if patients are left untreated. There are about 19,000 estimated new cases and 10,000 estimated deaths from this disease in 2018.[2,3]

*Corresponding author
Andy N.D. Nguyen, MD
Department of Pathology and Laboratory Medicine
University of Texas Health Science Center at Houston
6431 Fannin Street, MSB 2.292, Houston, TX, 77030
(713) 500-5337; fax (713) 500-0712
Nghia.D.Nguyen@uth.tmc.edu

There is an urgent need to find better treatments for this type of leukemia as only a quarter of the patients diagnosed with AML survive more than 5 years. AML includes many subtypes that share a common clinical presentation despite different types of mutations and genetic events. A variety of technologies targeting the gene, mRNA, microRNA and protein level have helped predicting the prognosis of AML patients. Interestingly, most AMLs only have only a few gene mutations, but prognosis of AML patients is quite varied. A possible explanation for this diversity is differences in protein signaling. The genetic aberrations and mutations of myeloid leukemic cells often cause a profound impact on the cellular protein networks.

Previous studies on the association between prognosis of AML and a small number of cytogenetic abnormalities and mutations highlighted the clinical and biologic heterogeneity of AML.[4-8] The cytogenetic abnormalities with prognostic relevance have led to the adoption of a risk stratification model: three cytogenetically defined risk groups with significant differences in overall survival.[9] Although risk stratification for AML patients has been improved to a great extent recently, a substantial number of patients still lack clear correlation between any specific abnormalities and accurate prognostic prediction. More recently, mutational analysis of FLT3, NPM1, and CEBPA was shown to improve risk stratification for AML patients without karyotypic abnormalities.[10] Recent advances in molecular studies, especially NGS, have identified additional recurrent somatic mutations in patients with AML, including mutations in TET2,[11-12] IDH1 and IDH2,[13-15] DNMT3A,[7,16] and PHF6,[17] among others. Retrospective studies indicate that a subset of these mutations may have prognostic significance in AML,[7,16,18] and these mutations may be the "missing" parameters in previous risk stratification models for patients with AML. Whether including these novel mutations in mutational profiling with a larger set of genes would improve prognostication of AML has not been investigated in clinical studies. Numerous mutations in AML have been found with recent application of NGS technique. The TCGA study[19] with 200 AML cases showed that the average number of mutations is 13 per case and there are 23 recurrent mutations. Such a large number of mutations in AML would certainly present a challenge in predicting prognosis for AML patients using multiple-variable statistical analysis.

In this paper, we propose the use of DL methods based on unsupervised feature extraction to address the challenge described above. Most successful DL methods involve artificial neural networks, a family of models inspired by biological neural networks (i.e. the central nervous system, particularly the brain). In such an artificial neural network, artificial nodes (known as "neurons") are connected to form a network mimicking a biological neural network. Warren McCulloch and Walter Pitts created a computational model for neural networks based on an algorithm called threshold logic in 1943.[20] Neural networks had not shown superior performance compared to other machine learning methods until the introduction of DL in 2006. DL is different from traditional machine learning in how representations are learned from the raw data. In fact, DL allows computational models that are composed of multiple processing layers based on neural networks to learn representations of data with multiple levels of abstraction.[21] Every layer of a deep learning system produces a representation of the observed patterns based on the data it receives as inputs from the previous layer, by optimizing a local unsupervised criterion.[22] The key aspect of deep learning is that these representations are not designed by human engineers, but they are learned from data using a general purpose learning procedure. Deep learning has recently shown impressive results in discovering intricate structures in high-dimensional data and obtained remarkable performances for object detection in images,[23,34] speech recognition,[25] natural language understanding[26] and translation.[27] Relevant clinical-ready successes have been obtained in health care as well (e.g. identification of metastatic breast cancer on lymph node slides,[28] aggregating features relevant to specific breast cancer subtypes,[29] predicting drug therapeutic uses and indications[30]), initiating the way toward a potential new generation of intelligent tools based on DL for real-world medical care.

We use stacked autoencoders which form a deep network capable of achieving unsupervised learning, a type of machine-learning algorithm which draws inferences from the input data and does not use labeled training examples. In contrast to previous methods of conventional neural network where data must be strictly categorized to provide the appropriate label for supervised learning, the unlabeled data in DL can be used in unsupervised training phase. The resulting features from all training sets are then used as the basis for constructing the classifier.

In this study, we use data from the TCGA database[19] which consist of 200 de novo AML cases and attempt to use DL which incorporates unsupervised feature training to find correlation between cytogenetics, age, mutation and prognosis. To the best of our knowledge, unsupervised feature learning methods has never been applied to predict AML prognosis in this manner.

**MATERIALS AND METHODS**
**Materials:**
Data from 200 cases of de novo AML were retrieved from TCGA database (public domain).[17] Demographic information shows: age 55±16.1, white 89%, black 8%, others 3%; male 54%, female 46%; normal cytogenetics 47%. Molecular testing was performed on multiple platforms: Affymetrix U133 Plus 2, Illumina Infinium Human Methylation 450 BeadChip, and Affymetrix SNP Array 6.0. All karyotypes were analyzed by conventional G-banding in at least 20 metaphases. Results were available for cytogenetics, 260 gene mutations, and survival duration (DTD) for each case.[31] As previously reported, in this database a total of 23 genes were significantly mutated, and another 237 were mutated in two or more samples.[31] Nearly all samples had at least 1 nonsynonymous mutation. To use the most relevant data for analysis, only cases with the following 23 most common mutations (grouped according to categories) were extracted for our study:
-Activated signaling (signal transduction): FLT3-ITD, KIT, KRAS, NRAS, and PTPN11
-Myeloid transcription factors (differentiation): NPM1, CEBPA, and RUNX1
-Epigenetic regulation: DNMT3A, TET2, IDH2, IDH1, EZH2, and HNRNPK
-Tumor suppressors: TP53, WT1, and PHF6
-Spliceosomes: U2AF1
-Cohesins: SMC1A, SMC3, STAG2, and RAD21,

-Non-annotated: FAM5C (BRINP3)

Furthermore, the following 10 common cytogenetic abnormalities were seen in the patient cohort: t(8;21), inv(16), t(15;17), t(9;11), t(9;22), trisomy 8, del (7), del (5), del (20), and complex chromosomal abnormalities. Subsequently only 94 cases with one or more of the 23 common mutations were selected and included in this study. These include cases with or without cytogenetic abnormalities. DTD was chosen as the prognostic parameter. For the 94 AML cases in this study, the mean DTD was 730 days. In summary, the number of input parameters was 34 (10 cytogenetics, age, and 23 mutations) and the number of outcome parameter (DTD) was one.

**Methods**

Our main analysis method was a DL neural network with stacked (multi-layered) auto-encoder. Training was mostly based on unsupervised feature learning which has been used successfully for image and audio recognition.[32,33] Our DL neural network was designed with the R language. R is a programming language for statistical computing and graphics supported by the R Foundation for Statistical Computing.[34] R was derived from the S language which was originally developed at Bell Laboratories by John Chambers and colleagues. R's popularity has increased substantially in recent years with advances in machine learning.[35] The source code for the R software environment is written primarily in Java, C, FORTRAN, and also in R itself. R is freely available under the GNU General Public License, and pre-compiled binary versions are provided for various operating systems including UNIX, Windows and MacOS. In this study, we used many DL functions obtained from various R packages which are available from the Comprehensive R Archive Network.[36]

The stacked autoencoder neural network, illustrated in Fig. 1, incorporates two training phases: pre-training with unsupervised learning method, and fine-tuning which is similar to the supervised back-propagation in conventional neural network.[37,38] During pre-training phase, the output from one layer is subsequently used as the input for the next output layer. The output from each layer essentially represents an approximation of the input data constructed from a limited number of features learnt by the hidden units of the network. The stacked autoencoder is constructed by multiple layers in the neural network (i.e. input layer, hidden layers, and output layer). For simplicity, only 2 layers are illustrated in Fig. 1. The sigmoid function is used as activation function in hidden layers. In fine-tuning phase, the back-propagation method minimizes the error with an additional sparsity penalty.[39] The features learned in the pre-training phase are subsequently used with a set of labeled data for specific status (positive or negative) to train a classifier. A classifier can be defined as a function that receives values of various features from training examples (cytogenetics, age, and mutations as independent variables) and provides an output which predicts the category that each training example belongs to (prognosis or DTD as dependent variable).[40] For the fine-tuning phase, we used linear function for the classifier.

Fig 1. A Deep Learning Neural Network (Stacked Autoencoder Network) with Unsupervised Training in Pre-training Phase and Supervised Training in the Fine-tuning phase

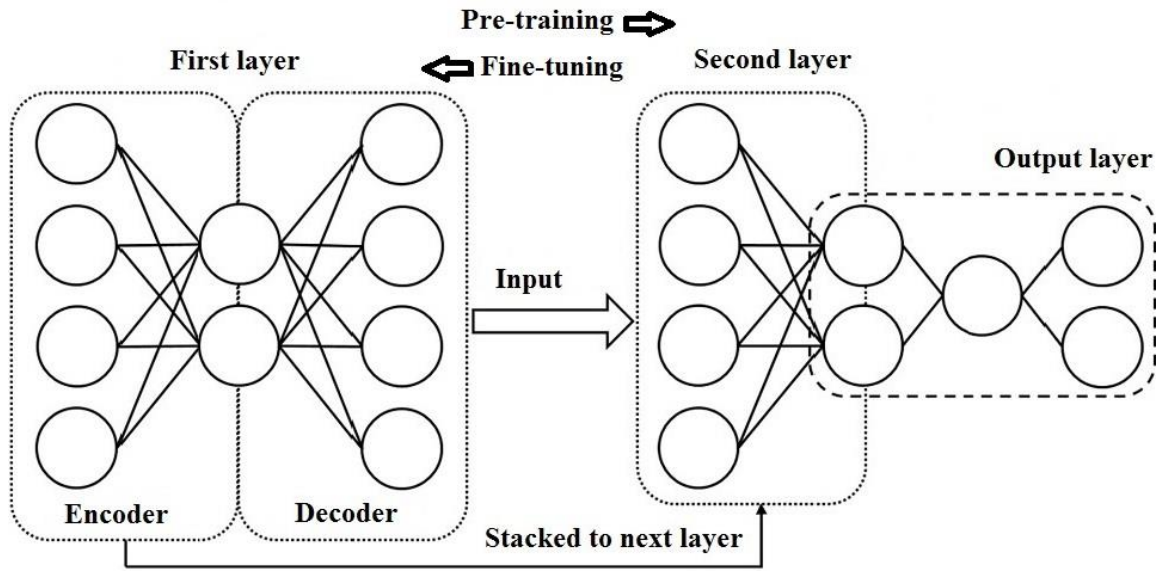

For the initial analysis, all 34 attributes (10 cytogenetics findings, age, and 23 mutations) were used to train the network to predict prognosis (good prognosis vs. poor prognosis, i.e. DTD is either more than or less than 730 days). A tenfold cross-validation method was used to obtain comprehensive validation results due to the small number of samples (94 cases). In this validation, a small subset of data (10 out of 94 cases) was excluded each time for training; the resultant trained network would be used to predict the prognostic status for each case in the excluded subset (the remaining 84 cases). The process was repeated until all 94 cases in the data set had been validated. The overall accuracy of the neural network is the mean of those for all the validated subsets. Subsequent analyses, using trial and error with different number of input parameters, are expected to show the optimal input for the most accurate prediction.

**RESULTS**

The initial use of the full attribute set (10 common cytogenetic abnormalities, age, and 23 common mutations) yielded 81% accuracy for predicting good prognosis of an AML case (day-to-death $\geq$ 730 days). This accuracy corresponded to a sensitivity of 74% and a specificity of 86% in predicting good prognosis of an AML case. The initial analysis showed that the following 14 attributes rank highest in term of predicting power among the 34 attributes: age, 7 cytogenetic abnormalities [tri8, del5, del7, complex chromosomal abnormalities, t(8;21), inv(16), t(15;17)], and 6 mutations [FLT3, NPM1, TP53, DNMT3, KIT, CEBPA]. Using these top-ranked 14 attributes, the DL network subsequently achieved a slightly better accuracy of 83%, with a sensitivity of 80% and a specificity of 85%. The accuracy in predicting prognostic status with different attribute sets by DL network is summarized in Table 1. Number of attributes smaller or larger than 14 did not yield better accuracy (data not shown) indicating that 14 is the optimal number of attributes for this study. It appeared that fewer than 14 attributes contain insufficient data for prediction. Conversely, more than 14 attributes introduced background noise, compromising accuracy.

Table 1. Accuracy in Predicting Prognostic Status with Different Attribute Sets by the Deep Learning Network

|  | 34 Attribute Set | | 14 Attribute Set | |
|---|---|---|---|---|
| Conventional | Validation Set No. | Accuracy | Validation Set No. | Accuracy |
|  | 1 | 90% | 1 | 90% |
|  | 2 | 80% | 2 | 70% |
|  | 3 | 80% | 3 | 80% |
|  | 4 | 90% | 4 | 90% |
|  | 5 | 100% | 5 | 90% |
|  | 6 | 80% | 6 | 80% |
|  | 7 | 70% | 7 | 70% |
|  | 8 | 70% | 8 | 80% |
|  | 9 | 80% | 9 | 100% |
|  | 10 | 75% | 10 | 80% |
|  | Mean= | 81%* | Mean= | 83% ** |

Legends:

*   corresponding to sensitivity of 74%, and specificity of 86%

** corresponding to sensitivity of 80%, and specificity of 85%

The use of machine learning algorithms frequently involves careful tuning of network configuration and learning parameters. This tuning often requires experience, and sometimes brute-force search.[41] During network training, we have tried various configurations for the neural network to achieve optimal accuracy and noted that our DL network performed best with 3 hidden layers consisting of 20, 15, 10 nodes, respectively.

The optimal learning parameters for our neural network, obtained through trial and error, were as follow[42]: Learning rate: 1.0, Momentum: 1.0, batch size=10, sigmoid function for activation and output.

We also noted that the 3 general types of attributes (cytogenetics, age, or mutations) are almost equally important in predicting prognosis as expected. By separately leaving out cytogenetics, age, or mutations in the analysis, the accuracy for prognosis prediction degraded significantly to 67%, 61%, and 74%, respectively.

## DISCUSSION

DL algorithms are new and innovative tools of research in machine learning to extract complex data representations at high levels of abstraction. In fact, DL has been cited as one of the 10 breakthrough technologies in 2013 by MIT Technology Review.[43] The most important contribution of DL algorithms is to develop a hierarchical architecture of data, where higher-level features are defined in terms of lower-level features. The hierarchical learning architecture of DL algorithms is motivated by the biological structure of the primary sensorial areas of the neocortex in the human brain, which automatically extracts abstract features from the underlying data.[44-46] DL algorithms rely on large amounts of unsupervised data, and typically learn data representations in a greedy layer-wise fashion.[47,48] Studies have shown that data representations obtained from stacking up nonlinear feature extractors (such as autoencoders used in our study) often yield better machine classification results.[49-51]

DL applications have produced outstanding results in several areas, including speech recognition,[52-56] computer vision,[47,48,57] and natural language processing.[58,59, 28] A recent challenge hosted by the International Symposium on Biomedical Imaging (ISBI) in 2016 lead to a successful DL system for automated detection of metastatic cancer from whole slide images of sentinel lymph nodes.[60] Data-intensive technologies as well as improved computational and data storage resources have contributed to Big Data science.[61] Technology-based companies such as Microsoft, Google, Yahoo, and Amazon have maintained databases that are measured in exabyte proportions or larger. Various private and public organizations have invested in Big Data Analytics to address their needs in business and research,[62] making this an exciting area of data science research.

In the present study, we used DL to predict prognosis of AML. Specifically, we rely on a set of attributes (cytogenetics, age, and mutations) to predict prognostic status in newly-diagnosed AML patients.  We implemented a DL network consisting of autoencoders that are stacked to form hierarchical deep models from which high-level features are compressed, organized, and extracted, without labeled training data. We showed how DL, which incorporates unsupervised feature training, can be used to predict prognosis using cytogenetics, age, and mutations with excellent results (accuracy of 83%, sensitivity of 80%, and specificity of 85%).

The main limitation of our preliminary study was the relatively small size of cohorts (94 cases out of 200 from TCGA database). Nevertheless, this study provided excellent preliminary results for future studies that may include many more cases, more cytogenetics and mutation data, and other clinical data such as co-morbidity index. With more data, it is expected that the accuracy would be higher than that from this preliminary study.

## CONCLUSION

DL method, a disruptive technology, is predicted to be an integrated part of future practice in molecular diagnosis and prognostic prediction using NGS data. Our preliminary study demonstrated a practical application in this area. The successful validation of this DL software is of tremendous value to personalized treatment of AML patients, i.e. stratifying treatment especially bone marrow/stem cell transplant for each patient based on predicted prognosis. The software's database can be continually kept up-to-date by adding new patients' data (with more patients, with additional tests, etc.) to improve its predicting ability. Furthermore, input ranking techniques in neural net can detect critical parameters which impact prognosis, and this helps to identify sets of important data to predict prognosis (novel patterns). While the amount of data used here was relatively modest, this study provided a proof-of-concept for using DL network as a more accurate approach for modeling big data in cancer genomics.

Acknowledgement: The data in this study were provided by TCGA database which is available in public domain for research purpose.